\documentclass{article}

\usepackage{graphicx} %

\usepackage{algorithm}

\usepackage{algpseudocode}

\usepackage{amsmath} %

\usepackage{amssymb}  %

\usepackage{url}

\usepackage{multirow} 

\usepackage{booktabs}

\usepackage{siunitx}

\newcommand{\bs}[1]{{\boldsymbol{#1}}}

\newcommand{\defeq}{\circeq}

\title{COAST (arXiv)}

\title{\LARGE \bf

COAST: Safe Flight Corridor Planner for Fast and Robust Snap Minimized Trajectories

}

\author{Alejandro Pelaez$^{1}$ and Randal W. Beard$^{1}$%
\date{November 2026}
\thanks{$^{1}$Multiple Agent Intelligent Coordination and Control (MAGICC) Laboratory, Brigham Young University, Provo, UT, USA. e-mail: {\tt\small \{apd915, beard\}@byu.edu}}%
}

\begin{document}

\maketitle

\thispagestyle{empty}

\pagestyle{empty}

\begin{abstract}
Current B-Spline trajectory planners face a critical trade-off: they either sacrifice computational speed to maintain strict Safe Flight Corridor (SFC) guarantees, or they abandon hard safety bounds in favor of soft-constraint penalties to achieve real-time performance. We present COAST (Corridor-Optimized Aerially-Safe Trajectories), a B-Spline based, online trajectory planner capable of formulating snap minimized trajectories with SFC guarantees. COAST divides the problem into two stages: an A* based front-end that generates geometrically simple SFCs (represented as a chain of rectangular prisms), and a quadratic programming back-end for control point optimization. To handle the non-convexity of SFC intersections, we leverage the finite support property of B-Splines, to ensure the spline never exits SFC boundaries. The control point optimization incorporates MINVO transforms for kinodynamic bounding, allowing COAST to design non-conservative, dynamically feasible trajectories, while a snap minimization matrix calculation provides flight smoothness. This formulation provides a formal geometric containment guarantee, while ensuring mean trajectory computation times under 5 ms for real-time deployment. We validate our results through high-fidelity simulation experiments. The source code can be found at: \url{https://github.com/byu-magicc/Minimum-Snap-Trajectories}
\end{abstract}

\section{INTRODUCTION}
 \begin{figure}[!htbp]
  \centering
  \includegraphics[width=0.6\linewidth]{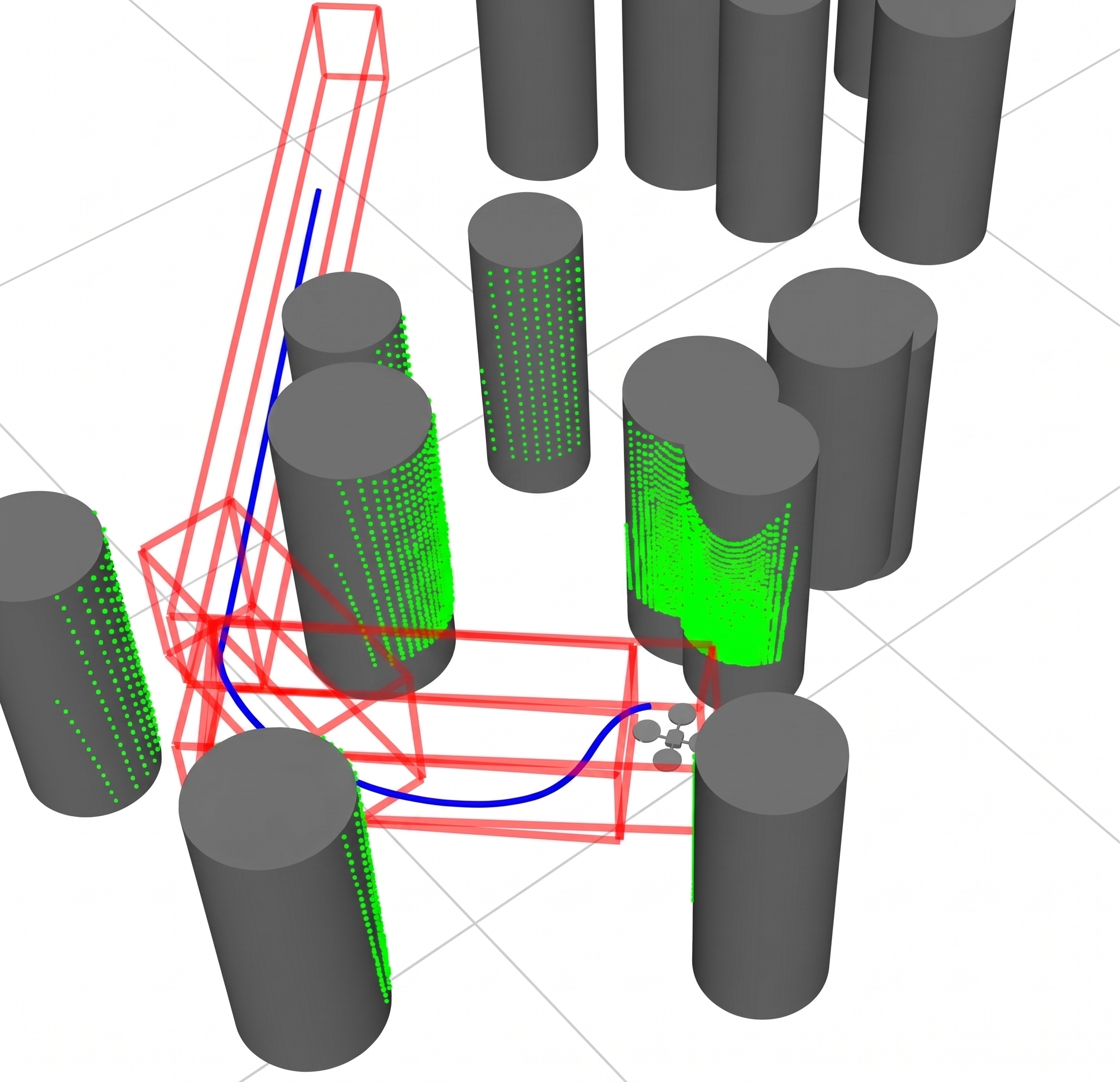}
  \caption{Application of COAST in a dense environment. The system utilizes LiDAR point cloud information (green) to design safe flight corridors (red), represented as a chain of dynamically-sizing rectangular prisms. A minimum snap, B-Spline based trajectory (blue) is then optimized within the corridors using MINVO transforms to account for kinodynamic limitations.}
  \label{fig:demo}
\end{figure} 

As expectations for fully autonomous aerial systems increase, real-time, or online, trajectory planning has become a critical research area, due to its direct contributions towards allowing systems to traverse unknown environments autonomously. In order for an online planner to be considered safe for real world deployment, it must know how to robustly address the multiple challenges that come from using only onboard data as the basis for decision making. Previous work varies in how different challenges, from approaches to obstacle avoidance, to how to represent the flight trajectory and make it dynamically feasible, are handled.

Obstacle avoidance has been a highly researched topic within trajectory planning, as a robust online planner must do an effective job in receiving and interpreting obstacle data in order to create a safe trajectory in computationally reasonable times. As such, different methods for obstacle avoidance have been proposed. References~\cite{zhou2019robust, zhou2020ego} propose penalty-based methods, where collisions within a trajectory are not prohibited but heavily penalized, leading to a lack of formal guarantees of collision-free trajectories in favor of computationally fast planning times. References~\cite{kumarsfc, tetrahedron} propose safe flight corridors (SFCs) as a method to ensure obstacle-free trajectories. SFC-based trajectory planning in turn trades off computational speed, for more robust guarantees of collision-free trajectories due to the involvement of volumes for the aircraft to plan trajectories through, that are mathematically proven to be devoid of obstacles.

Multiple curve representations have been used to design the trajectory. Reference~\cite{mighty} uses hermite splines, with \cite{sando} choosing bezier curves as a representation. B-Splines are a widely used standard for trajectory representation, with \cite{mader, zhou2019robust, zhou2020ego, corbplanner, tetrahedron, RRTPlanner} using them as their method of choice. For trajectory smoothness \cite{kumarsfc} and~\cite{Richter2016} minimize snap as part of their formulation, while others such as \cite{MuellerJerk} minimize jerk.

COAST aims to expand on SFC-bounded B-Spline trajectories, as done in \cite{corbplanner, tetrahedron}. However, our general contributions, which span more efficient SFC formulation methods and snap minimization using vector representation of B-Splines for faster optimization, are as follows:
\begin{enumerate} 
\item A snap minimization method using vector representations of B-Splines to formulate a quadratic cost function dependent on control point placement.
\item A rectangular prism SFC formulation method adjusted for multirotor flight building on the representation for fixed-wing flight from \cite{RRTPlanner}.
\item A coupling QP optimization method that minimizes snap as constrained by SFC boundaries and kinodynamic limitations. 
\end{enumerate}

\section{B-SPLINES}
This paper uses a $4^{th}$ degree clamped B-Spline representation for trajectory formulation, with uniform knot spacing.  

B-Spline trajectories are defined as 
\[
\boldsymbol{p}(t) = \sum_{m=0}^{M+d-1} \boldsymbol{c}_m \hat{b}_m^d(t, \boldsymbol{k}),
\]
where $\boldsymbol{c}_m \in \mathbb{R}^n$, $m = 0, \dots, M+d-1$ are the control points, $\boldsymbol{k} = (\tau_0, \tau_1, \tau_2, \dots, \tau_T)$ is the knot sequence, and $b_m^d(t, \boldsymbol{k})$ are the B-spline basis functions of degree $d$, given the knot sequence $\boldsymbol{k}$. We decide whether a B-Spline is categorized as natural or clamped based on its knot sequence.

\subsection{Clamped and Uniform Knot Sequences}
The sequence 
\[
\boldsymbol{k}_M^d = \big[ \underbrace{t_{0}, \dots, t_{0}}_{d\text{-terms}}, \, \underbrace{t_0, t_1, \dots, t_M}_{(M+1)\text{-terms}}, \, \underbrace{t_{M}, \dots, t_{M}}_{d\text{-terms}} \big]
\]
is \emph{clamped} if $M > d$ and its middle $(M+1)$ terms are strictly increasing, where $t_i < t_{i+1}$ for $i = 0, \dots, M-1$, with $t_0$ and $t_M$ each appearing $d+1$ times at their respective ends. 

It is additionally \emph{uniform} if the middle $(M+1)$ terms are equally spaced, i.e. for $\Delta > 0$ it has form 
\[
\underbrace{t_0, t_0 + \Delta, t_0 + 2\Delta, \dots, t_0 + M\Delta}_{(M+1)\text{-terms}}.
\]

A \emph{natural} sequence, by contrast, leaves the leading and trailing $d$ knots distinct, which forces boundary control points outside the trajectory endpoints as shown in Fig.~\ref{fig:clamped_vs_natural}. 

\subsection{Finite Support \& Convex Hull Properties}
The finite support and convex hull properties are two of the most critical properties of B-splines for trajectory generation.

The finite support property, alternatively known as local control, dictates that at any given instant of time $t \in [t_s, t_{s+1})\subset[0,M]$, there are exactly $d+1$ non-zero basis functions. Consequently, the trajectory $\boldsymbol{p}(t)$ over this interval is only influenced by the local subset of $d+1$ control points, namely $\{\boldsymbol{c}_s, \dots, \boldsymbol{c}_{s+d}\}$. This finite support ensures that modifying a single control point only alters the trajectory locally.

The convex hull property, making use of finite support, guarantees that, because the active B-spline basis functions are non-negative and sum to unity, the trajectory segment $\boldsymbol{p}(t)$ for $t \in [t_s, t_{s+1})\subset[0,M]$ is strictly contained within the convex hull of those active supporting control points, i.e., 
\[
t\in[s,s+1] \implies \boldsymbol{p}(t) \in \text{conv}\{\boldsymbol{c}_s, \dots, \boldsymbol{c}_{s+d}\}.
\]
This property bridges continuous trajectory planning with discrete spatial bounding: ensuring that the local control points remain within a collision-free convex region (such as a Safe Flight Corridor), mathematically guarantees that the continuous curve is also collision-free.

\subsection{MINVO Bounds}
 \begin{figure}[!htbp]
  \centering
  \includegraphics[width=0.6\linewidth]{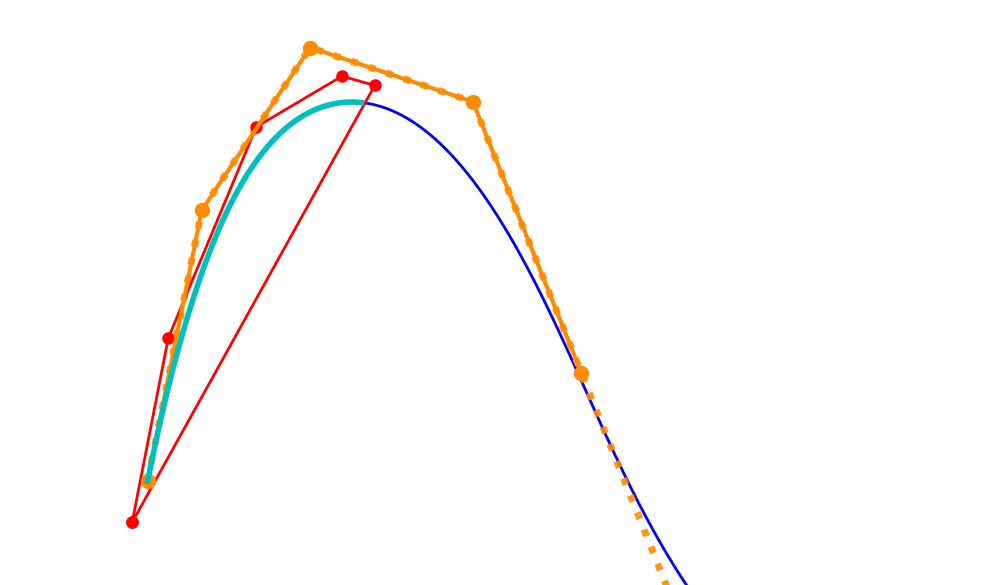}
  \caption{B-Spline (orange) vs MINVO (red) clamped control points comparison. Each $j$ interval of the MINVO control points shifts with each equivalent set of $d+1$ B-Spline control points. MINVO control points create a tighter convex hull bound around the spline trajectory.}
  \label{fig:minvo}
\end{figure} 
The work presented by \cite{tordesillas2022minvo} describes MINVO bounds as a series of matrix transformations applied to B-Spline control points. This transformation creates a \textbf{\underline{min}}imum \textbf{\underline{vo}}lume control point representation for the spline trajectory as shown in Fig.~\ref{fig:minvo}. A MINVO representation is defined and calculated by \cite{mader} as follows:
\begin{equation}
    \{\boldsymbol{v}_s, \dots, \boldsymbol{v}_{s+d}\}^{\text{MV}} = \{\boldsymbol{c}_s, \dots, \boldsymbol{c}_{s+d}\}^{\text{BS}} \boldsymbol{A}^{\text{BS}}_d \left( \boldsymbol{A}^{\text{MV}} \right)^{-1}
\end{equation}
where $\{\boldsymbol{v}_s, \dots, \boldsymbol{v}_{s+d}\}^{\text{MV}}$ are the MINVO control points for $t\in[s,s+1]$. The matrix $\boldsymbol{A}^{\text{MV}}$ can be found in \cite{tordesillas2022minvo}, while $\boldsymbol{A}^{\text{BS}}_d$ can be computed based on the work in \cite{Qin1998}. 

By applying this MINVO transformation to the trajectories and their derivatives, namely velocity and acceleration, we obtain tight bounds, which are critical for maximizing aircraft performance within our framework.

\section{METHODS}

\subsection{Snap Minimization with B-Splines}
\label{sec:snap-minimization}
The work in \cite{kumarsnap} demonstrates that since acceleration is dependent on the second derivative of position, smooth trajectories for quadrotors can be achieved by minimizing the fourth derivative of position (snap). We minimize snap by using a vector representation of B-Splines to create penalization matrices to constrain control point placement. 

We define a B-Spline in vector form as
\[
\bs{p}(t) = \bs{C} \bs{b}_M^d(t, \bs{k}_M^d), \qquad t \in \text{span}(\bs{k}_M^d),
\]
where
\[
\boldsymbol{C} \doteq \begin{bmatrix} \boldsymbol{c}_0 & \dots & \boldsymbol{c}_{M+d-1} \end{bmatrix},
\]
\[
\bs{b}_M^d(t, \bs{k}_M^d) \defeq \begin{pmatrix} b_0^d(t, \bs{k}_M^d) \\ b_1^d(t, \bs{k}_M^d) \\ \vdots \\ b_{M+d-1}^d(t, \bs{k}_M^d) \end{pmatrix}.
\]
A useful property of B-splines is that their derivatives yield another B-spline of degree $d-1$. Furthermore, the control points of this derivative spline are given by the scaled differences of adjacent original control points (i.e., $\boldsymbol{c}_{m+1} - \boldsymbol{c}_m$). The scaling factor for each difference depends on the local knot span, which properly accounts for the knot multiplicities at the boundaries of a clamped spline. Because this scaled differencing is a linear operation, we can compute the derivative in vector form using a sparse transformation matrix: 
\[
\frac{d\boldsymbol{p}}{dt}(t) = \boldsymbol{C} D_M^d \boldsymbol{b}_M^{d-1}(t),
\]
where $D_M^d$ is the $(M+d) \times (M+d-1)$ derivative matrix defined as: 
\[
D_M^d = -\begin{bmatrix}\bar{D}_M^d \\ \mathbf{0}_{1\times(M+d-1)} \end{bmatrix} + \begin{bmatrix}\mathbf{0}_{1\times(M+d-1)} \\ \bar{D}_M^d \end{bmatrix},
\]
and
\[
\bar{D}_M^d = \text{diag}\Big(\tfrac{d}{1}, \tfrac{d}{2}, \dots, \tfrac{d}{d-1}, \underbrace{1, \dots, 1}_{M-d}, \tfrac{d}{d-1}, \dots, \tfrac{d}{2}, \tfrac{d}{1}\Big).
\]
Because this differencing operation is recursive, higher-order derivatives such as snap, can be computed by stacking successive derivative matrices. To simplify notation for the $\ell$-th derivative (where $\ell = 1$ for velocity, $\ell = 4$ for snap), we define the higher-order derivative operator as the product of these matrices: 
\[
D_M^{d, \ell} \defeq D_M^d D_M^{d-1} \dots D_M^{d-\ell}.
\]
To generate smooth trajectories, we seek to find the optimal control points $\boldsymbol{C}$ that minimize the integral of the squared norm of the trajectory's derivatives (where penalizing the $4^{th}$ derivative minimizes snap). By substituting our matrix formulation into this integral, the objective function reduces to a purely algebraic form.

First, we define the integral of the outer product of the basis vectors for an arbitrary degree $q$ over the curve duration as 
\[
S_M^q \defeq \int_{0}^{M} \mathbf{b}_M^q(t, \bs{k}_M^q) \mathbf{b}_M^q(t, \bs{k}_M^q)^\top dt.
\]
Using this pre-computable integral, the penalization matrix for the $\ell$-th derivative is constructed by wrapping it in our derivative operator matrices: 
\[
W_M^{d, \ell} \defeq D_M^{d, \ell-1} S_M^{d-\ell} {D_M^{d, \ell-1}}^\top.
\]
Ultimately, the complex integral objective of minimizing the trajectory's derivatives elegantly reduces to a convex quadratic cost function solely dependent on the control points: 
\begin{equation}
\label{eq:snap_min}
    J(\boldsymbol{C}) = \text{trace} \left( \boldsymbol{C} W_M^{d, \ell} \boldsymbol{C}^\top \right).
\end{equation} 
Because $W_M^{d, \ell}$ is constant and completely independent of the control points, it can be pre-computed offline. This formulation allows the snap minimization problem to be solved extremely efficiently as a standard Quadratic Program (QP).

\subsection{Safe Flight Corridors (SFCs)}
We begin our Safe Flight Corridor (SFC) formulation by using an A* algorithm to find an initial feasible path over a voxelized world-map representation. To reduce the number of SFCs and to smooth the path, we use a greedy algorithm to find the first node, beginning at the start node, where a straight-line ray collides with an obstacle. We then select the preceding node, as it was proven to be collision-free, and prune all intermediate nodes leaving only the starting node and the collision-free node. We iterate, using the last collision-free node as the new start node, until we reach the end point. This algorithm has been successfully used to find the minimum number of SFCs required for a given path, as shown in \cite{corbplanner}. 

While the SFCs are represented as a chain of rectangular prisms as originally outlined in \cite{RRTPlanner}, our implementation diverges through a true 3D application and by dynamically resizing corridor width and height according to the local obstacle environment. 
Algorithm~\ref{alg:dynamic_sfc} shows how the SFC bounds resize asymmetrically based on obstacle positioning. We base our formulation directly on the path segment origin and local geometric bounds. Let $p_A \in \mathbb{R}^3$ be the origin point of the $i$-th trajectory segment, and let $(\boldsymbol{u}_{i,x}, \boldsymbol{u}_{i,y}, \boldsymbol{u}_{i,z}) \in SO(3)$ be the local orthogonal basis vectors describing the spatial orientation of the corridor.

\begin{algorithm}[!htbp]
\caption{Dynamic 3D SFC Bounding via Spatial Projection}
\label{alg:dynamic_sfc}
\begin{algorithmic}[1]
\Require Segment origin $p_A \in \mathbb{R}^3$, Segment end $p_B \in \mathbb{R}^3$, KD-Tree $\mathcal{K}$
\State Calculate local orthogonal axes $u_x, u_y, u_z$ from vector $p_B - p_A$
\State Initialize bounding extents $e_i$ to maximum flight limits
\State $R \gets$ maximum bounding radius of the initial corridor
\State $O \gets$ \Call{QueryRadius}{$\mathcal{K}, p_A, p_B, R$}
\For{\textbf{each} point $p_{obs} \in O$}
    \State $v \gets p_{obs} - p_A$
    \State $P_x \gets v \cdot u_x$, $P_y \gets v \cdot u_y$, $P_z \gets v \cdot u_z$
    \If{projected distances $(P_x, P_y, P_z)$ lie within bounds $e_i$}
        \State Determine axis of maximum penetration
        \State Shrink corresponding bound in $e_i$ to exclude $p_{obs}$
    \EndIf
\EndFor
\State $A_i \gets \begin{bmatrix} u_{x} & -u_{x} & u_{y} & -u_{y} & u_{z} & -u_{z} \end{bmatrix}^\top$
\State $b_i \gets A_i p_A + e_i$
\State \Return $A_i, b_i$
\end{algorithmic}
\end{algorithm}

\begin{figure}[!htbp]
  \centering
  \includegraphics[width=0.6\linewidth]{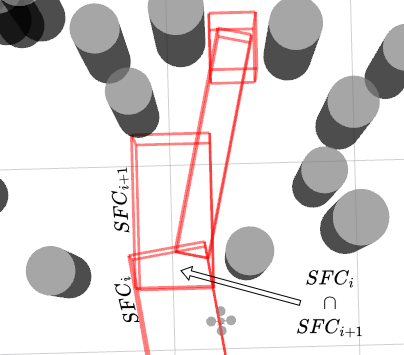}
  \caption{Dynamically-sizing SFCs. The obstacles and environment around the aircraft dictate the width and height of the individual rectangular prisms.}
  \label{fig:SFCs}
\end{figure}

We form a constraint matrix $A_i \in \mathbb{R}^{6 \times 3}$ containing the six outward-facing normal vectors as its rows:
\[
A_i = \begin{bmatrix} \boldsymbol{u}_{i,x} & -\boldsymbol{u}_{i,x} & \boldsymbol{u}_{i,y} & -\boldsymbol{u}_{i,y} & \boldsymbol{u}_{i,z} & -\boldsymbol{u}_{i,z} \end{bmatrix}^\top.
\]
Let $e_i \in \mathbb{R}^6$ denote the vector of maximum collision-free distances from $p_A$ along each corresponding normal direction, derived via spatial obstacle projections (e.g., Separating Axis Theorem). The offset vector $b_i \in \mathbb{R}^6$ for the half-planes is determined by projecting the segment origin onto the basis and adding the maximum extents:
\[
b_i = A_i p_A + e_i.
\]
A point $x \in \mathbb{R}^3$ resides in the safe flight corridor if it satisfies the element-wise linear inequality:
\[
A_i x \leq b_i.
\]
Fig.~\ref{fig:SFCs} demonstrates the dynamic width and height resizing along with the intersection areas of the SFCs.

\subsection{Trajectory Optimization}
 \begin{table*}[!htbp]
\centering
\caption{Benchmark Results From 1,000 Randomized Forests at 5\% Obstacle Density Without a Closed-Loop Controller. Timings Aggregate 103,687 Quadratic Program Solves.}
\label{tab:monte_carlo_offline}
\small
\setlength{\tabcolsep}{4pt}
\resizebox{\linewidth}{!}{%
\begin{tabular}{llcllcll}
\toprule
\multicolumn{2}{c}{\textbf{Mission Outcomes}} & &
\multicolumn{2}{c}{\textbf{Computational Timing} [ms]} & &
\multicolumn{2}{c}{\textbf{Trajectory \& Planner}} \\
\cmidrule(lr){1-2} \cmidrule(lr){4-5} \cmidrule(lr){7-8}
Success rate $R_{succ}$ [\%] & 81.9 & &
  Replan, mean       & 3.12      & &
  First-attempt solves [\%]  & 91.0 \\
Stalled [\%]                 & 17.8 & &
  Replan, med                & 1.76         & &
  Control points             & 8.66 $\pm$ 1.47 \\
Collided [\%]                & 0.3  & &
  OSQP, mean         & 0.67      & &
  Single-corridor plans [\%] & 59.4 \\
Median progress [\%]         & 97.5 & &
  OSQP, med                  & 0.32   & &
  Plan-mean speed [m/s]      & 2.59 $\pm$ 0.24 \\
\bottomrule
\end{tabular}%
}
\end{table*} 

The resulting flight tunnel generated by our front-end is a series of individually convex SFCs, rendered non-convex by their union. Traditional unconstrained minimum-snap analytical solvers are fundamentally incapable of enforcing the complex spatial inequalities required to keep a trajectory inside such non-convex bounds. Therefore, COAST formulates the trajectory generation as a constrained Quadratic Program (QP) and utilizes the Operator Splitting Quadratic Program (OSQP) solver \cite{Stellato_2020}.

OSQP leverages the Alternating Direction Method of Multipliers (ADMM) to rapidly solve problems in the standard quadratic form: 
\[
\begin{aligned}
\text{minimize} \quad & \frac{1}{2} \boldsymbol{x}^\top P \boldsymbol{x} + \boldsymbol{q}^\top \boldsymbol{x} \\
\text{subject to} \quad & \boldsymbol{l} \leq A \boldsymbol{x} \leq \boldsymbol{u}
\end{aligned}
\]
where $\boldsymbol{x} \in \mathbb{R}^{3N}$ represents the flattened, one-dimensional vector of all 3D position control points.

To map our minimum-snap objective to this standard solver format, we assign Eq.~\eqref{eq:snap_min} as the minimization function. The quadratic cost matrix $P$ is constructed by expanding the single-dimensional penalization matrix $W_M^{d, \ell}$ across the three spatial dimensions using a Kronecker product, while the state vector $\boldsymbol{x}$ represents the flattened, three-dimensional set of control points $\boldsymbol{C}$. Because our objective purely penalizes the squared derivatives of the curve without linear spatial costs, the linear cost vector $\boldsymbol{q}$ is strictly zero.

The true flexibility of this formulation lies in the sparse constraint matrix $A$. By vertically stacking distinct linear constraints, we guarantee both geometric safety and dynamic feasibility in a single, unified solve. The $A$ matrix is composed of four blocks, grouped below:

\noindent\textbf{Boundary Conditions ($A_{eq}$):} Strict equality constraints (where $\boldsymbol{l} = \boldsymbol{u}$) that precisely pin the initial and final position, velocity, and acceleration of the B-spline to the quadrotor's current state and target goal state.

\noindent\textbf{Spatial Safety ($A_{sfc}$):} Inequality constraints derived from the geometric half-planes of the SFCs. In order to safely navigate the non-convexity of the overall flight tunnel, we utilize the convex hull and finite-support properties of uniform B-splines. By strictly constraining subsets of sequential control points to lie within their respective convex SFCs (and specifically constraining shared control points to lie within the overlapping intersection of sequential SFCs), we mathematically guarantee that the entire continuous B-spline curve remains safely contained within the non-convex union. 

\begin{figure}[!htbp]
    \centering
    \includegraphics[width=0.65\linewidth]{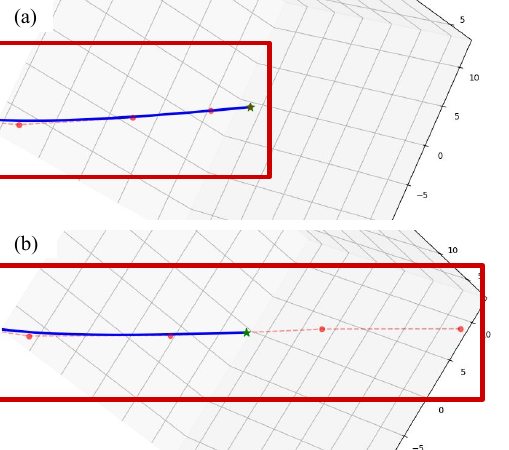}
    \caption{Impact of knot sequence selection on SFC bounding. (a) Clamped B-splines pin boundary control points directly to the trajectory endpoints. (b) Natural B-splines require control points outside the trajectory's physical endpoints.}
    \label{fig:clamped_vs_natural}
\end{figure}

As shown in Fig.~\ref{fig:clamped_vs_natural}, the use of clamped splines allows for more accurate SFC formulation, as natural splines require control points extending beyond the trajectory's endpoints. The starting and ending control points of a clamped B-Spline are tightly coupled with the trajectory's endpoints, saving valuable optimization time as they do not require precise spatial placement.

\noindent\textbf{Kinodynamic Limits ($A_{vel}, A_{accel}$):} To ensure the generated path does not exceed the physical velocity and thrust limits of the quadrotor, we must bound the trajectory's derivatives. As derived in Section~\ref{sec:snap-minimization}, the $\ell$-th derivative of the trajectory can be explicitly evaluated via the matrix operator $D_M^{d, \ell-1}$: 
\[
\boldsymbol{p}^{(\ell)}(t) = \boldsymbol{C} D_M^{d, \ell-1} \boldsymbol{b}_M^{d-\ell}(t).
\]
This matrix formulation allows us to express higher-order kinodynamic properties (like velocity and acceleration) as direct, linear transformations of the original position control points. By extracting these derivative control points into the $A$ matrix, we can enforce bounding constraints. However, rather than applying standard B-spline bounding---which yields overly conservative estimates---we apply the MINVO matrix transformations. MINVO control point representations allow us to get much more accurate and less conservative estimates on velocity and acceleration, allowing the solver to generate highly aggressive trajectories without violating physical vehicle constraints.

Because B-splines possess local support---meaning adjusting one control point only alters a small localized segment of the curve---the resulting $P$ and $A$ matrices are sparse. OSQP natively exploits this compressed sparsity, allowing COAST to solve the heavily constrained trajectory optimization in real-time.

\section{SIMULATION RESULTS}
 \begin{table*}[!htbp]
\centering
\caption{Kinodynamically-Enabled Closed-Loop Benchmark. Under Realistic Flight Conditions, COAST Generates Snap Minimized Trajectories With $<\SI{5}{ms}$ Mean Computation time and Strict Adherence to Per-Axis Kinodynamic Limits ($\rho_{vel} = \rho_{acc} = 0$).}
\label{tab:benchmark_results}
\setlength{\tabcolsep}{4pt}
\resizebox{\linewidth}{!}{%
\begin{tabular}{l c c c c c c c c}
\toprule
\multirow{2}{*}{\textbf{Density}} & \textbf{Success} & \multicolumn{2}{c}{\textbf{Comp. Time}}
  & \multicolumn{3}{c}{\textbf{Performance}} & \multicolumn{2}{c}{\textbf{Constraint Viol.}} \\
\cmidrule(lr){2-2} \cmidrule(lr){3-4} \cmidrule(lr){5-7} \cmidrule(lr){8-9}
 & $R_{succ}$ [\%] & $T_{opt}$ [ms] & $T_{replan}$ [ms]
 & $T_{trav}$ [s] & $L_{path}$ [m] & $S_{jerk}$ [m/s$^2$]
 & $\rho_{vel}$ [\%] & $\rho_{acc}$ [\%] \\
\midrule
5\% & 100.0 & 0.89 & 2.46 & 47.48 & 109.79 & 64.58 & 0.00 & 0.00 \\
10\% & 90.0 & 1.10 & 4.25 & 50.16 & 119.68 & 129.20 & 0.00 & 0.00 \\
\bottomrule
\end{tabular}%
}
\end{table*} 

We performed all simulations on a desktop computer equipped with an Intel\textregistered\ Core\texttrademark\ i7-7700 CPU @ 3.60GHz and 16 GB of RAM. The operating system is Ubuntu 24.04 LTS, and we utilized ROS 2 Jazzy for the implementation. To evaluate COAST, we use the methodology established in \cite{sando}. This framework generates randomized and static forest environments populated with cylindrical obstacles at varying density levels, namely 5\% and 10\%, in a $100 \times \SI{40}{m}$ field, traversed from $(0,0,3)$ to $(105,0,3)$. Cruise altitude is held at \SI{3}{m} against \SI{6}{m} trunks so that climbing over the obstacle field is not an admissible shortcut, and every environment was verified prior to testing to admit a collision-free route at an inflation exceeding the \SI{0.69}{m} footprint used by the planner. 

We define the success rate ($R_{succ}$) as a trial where the vehicle reaches the goal while registering zero collisions. Velocity and acceleration limits were fixed at $v_{max} = \SI{5}{m/s}$ and $a_{max} = \SI{20}{m/s^2}$ for all simulation runs, with a violation of either limit being penalized. Our evaluation is divided into two phases: first, an extensive offline Monte Carlo analysis to validate the algorithm's raw computational performance and mathematical safety bounds; and second, a fully closed-loop simulation using the ROSflight framework to verify the dynamic feasibility of the generated trajectories.

\subsection{Monte Carlo Benchmarking}
Table~\ref{tab:monte_carlo_offline} reports 1,000 independent trials conducted without a closed-loop controller, isolating planner performance from tracking error. Any failure is therefore attributable to the planner rather than to an infeasible map.

Across 103,687 quadratic program solves, COAST sustains a median solve time of \SI{0.32}{ms}, and 91.0\% of trajectories are recovered on the first solve attempt without time-budget relaxation. The median plan requires only eight control points (mean 8.66), confirming the formulation stays compact in clutter. 

Of the 18.1\% of trials not reaching the goal, stalls outnumber collisions by roughly 59 to 1, as contact occurs in 0.3\% of trials, while the median trial closes 97.5\% of the start-to-goal distance. Hard corridor constraints render infeasible paths a solver failure rather than a safety violation, which is the central practical advantage of SFC bounding over penalty-based avoidance.

\subsection{ROSflight Benchmarking}
To validate the dynamic feasibility of the generated trajectories, the quadrotor physics and low-level control dynamics were simulated using the open-source ROSflight framework \cite{rosflight}, while ROScopter \cite{roscopter} was utilized as the high-level multirotor autopilot to execute trajectory tracking.

To analyze safety and feasibility, we track $\rho_{vel}$ and $\rho_{acc}$, which represent the fraction of planned commanded setpoints violating the per-axis ($L_\infty$) limits. Because jerk is not explicitly constrained by our QP formulation, it is reported purely as a performance metric ($S_{jerk}$) rather than a violation rate. We record computation times as $T_{opt}$ for the QP optimization alone and $T_{replan}$, which includes path planning, matrix construction, and QP optimization.

Table~\ref{tab:benchmark_results} reports ten independent trials at each density. Most significantly, $\rho_{vel} = \rho_{acc} = 0$ for every trial at both densities: not one commanded setpoint violates the per-axis velocity or acceleration limit. This validates the MINVO bounding formulation end-to-end, since the constraints are imposed on control points at optimization time yet are shown to hold on the densely sampled trajectory actually executed. COAST reaches the goal without contact in all ten trials at 5\% density and nine of ten at 10\%, with the singular failure being a stall.

End-to-end replanning averages \SI{2.46}{ms} and \SI{4.25}{ms} at 5\% and 10\% density respectively. The success rate exceeds that of the open-loop study because the closed-loop system continues executing its committed setpoint buffer through a failed replan, whereas the offline rollout advances only along each newly solved trajectory. The median control-point count remains at eight in all but one trial across both densities, which also bounds achievable speed: the snap cost stencil requires four spline segments, so every trajectory spans at least eight control points, and under uniform knots this imposes a floor on commanded duration independent of path length. The resulting conservatism is visible in $T_{trav}$ and motivates the non-uniform knot sequences discussed as future work in Section~\ref{sec:conclusions}.

\section{CONCLUSIONS}
\label{sec:conclusions}
This work introduced COAST, a safe, minimum snap trajectory planner for smooth flight in realistic scenarios. COAST formulates trajectory generation as a Quadratic Programming (QP) problem, where B-spline control points are optimized subject to strict start and goal states. By tightly constraining these control points within dynamically-sized Safe Flight Corridors (SFCs) and utilizing MINVO transforms to enforce kinodynamic limits, the solver inherently guarantees mathematical safety, while a penalty matrix minimizes snap subject to the constraints in place. In closed-loop simulations utilizing ROSflight and ROSCopter, COAST maintained zero observed violations in the densely sampled simulation trajectories while achieving mean end-to-end replanning times under 5 ms.

While COAST demonstrates robust real-time performance, the current framework exhibits limitations, including occasional flight stalling and conservative velocity profiles bounded by control point over-allocation and the uniform knot structure. Future work will focus on three primary objectives. First, we will explore advanced local and global horizon algorithms to mitigate flight stalling. Second, to safely increase flight velocities, we plan to evaluate non-uniform knot sequence integration into the trajectory optimizer through control point representation transformations. Finally, we aim to validate COAST through rigorous indoor and outdoor hardware experiments, ultimately extending this kinodynamically feasible architecture to fixed-wing platforms.

\bibliographystyle{IEEEtran}
\bibliography{main}

\end{document}